\documentclass[runningheads]{llncs}

\usepackage{eccv}

\usepackage{eccvabbrv}
    
\usepackage{graphicx}
\usepackage{caption}
\usepackage{booktabs}
\usepackage{multirow}
\usepackage{array}
\usepackage{subcaption}
\usepackage{wrapfig}
\usepackage{amsmath} 

\usepackage[accsupp]{axessibility}  % Improves PDF readability for those with disabilities.

\usepackage{hyperref}

\usepackage{orcidlink}

\begin{document}

\title{DMAConv: Dual Mask-Adaptive Convolution for Remote Sensing Pansharpening} 

\titlerunning{Dual Mask-Adaptive Convolution for Remote Sensing Pansharpening}

\author{
Xianghong Xiao\inst{1}\textsuperscript{\textdagger} \and
Zeyu Xia\inst{1}\textsuperscript{\textdagger} \and
Zhou Fei\inst{2} \and
Jinliang Xiao\inst{1} \and
Haorui Chen\inst{1} \and
Liangjian Deng\inst{1}\textsuperscript{*}
}
\authorrunning{X.~Xiao et al.}

\institute{University of Electronic Science and Technology of China, Chengdu 611731, China \and
Tongji University, Shanghai 200092, China}

\makeatletter
\def\blfootnote{\gdef\@thefnmark{}\@footnotetext}
\makeatother
\blfootnote{\textsuperscript{\textdagger}These authors contributed equally to this work.}
\blfootnote{\textsuperscript{*}Corresponding author.}

\maketitle

\begin{abstract}
  Pansharpening aims to fuse a high-resolution panchromatic image with a low-resolution multispectral image. Existing deep learning methods, including recent adaptive convolutions, struggle with regional heterogeneity in remote sensing images and often incur prohibitive computational costs. To address these challenges, we propose Dual Mask-Adaptive Convolution (DMAConv), a novel operator that dynamically allocates computational resources based on feature characteristics. DMAConv first employs a lightweight module to generate soft and hard masks. The hard mask separates features into a compact branch for processing redundant information globally and a focused branch that models complex, heterogeneous regions with greater computational investment. The soft mask then preliminarily modulates the input features for both branches. This dual-branch, mask-adaptive design significantly enhances feature representation while minimizing computational overhead. Extensive experiments demonstrate that our method achieves SOTA on a broad array of quantitative benchmarks, with substantially lower parameter counts and the minimal computational cost among adaptive convolution models.
\end{abstract}

\begin{figure}[t]
    \centering
    \includegraphics[width=1\linewidth]{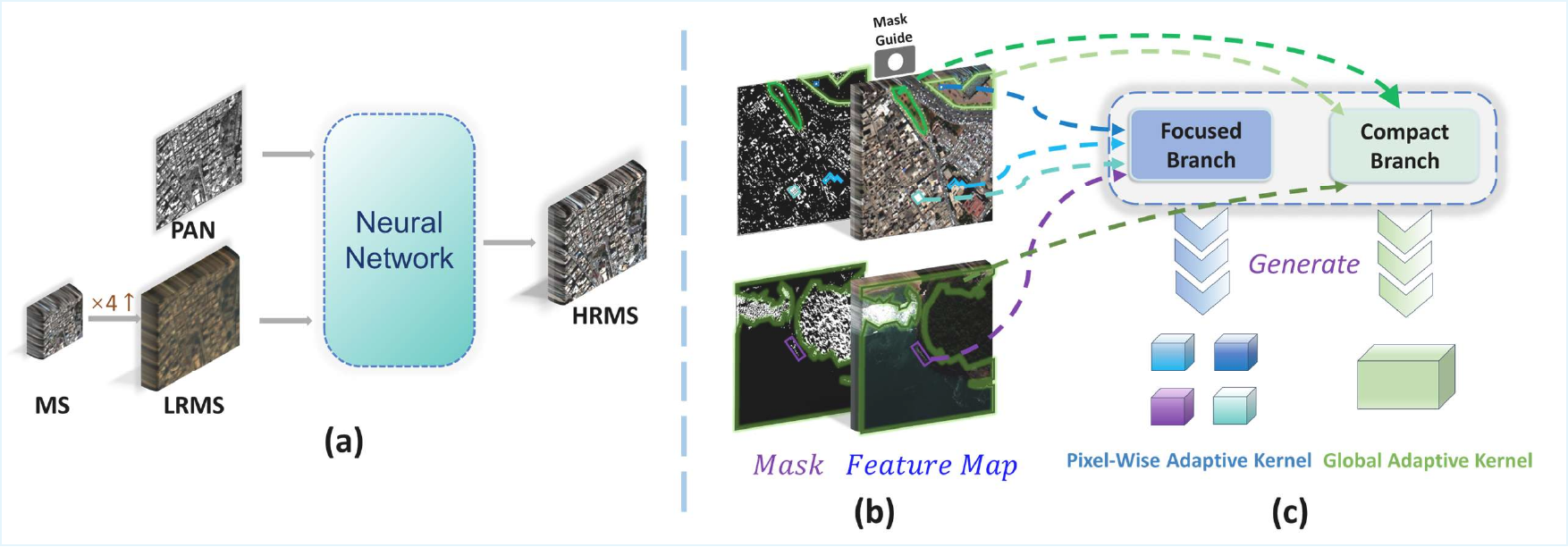}
    \caption{\textbf{(a)} Conventional deep learning–based pansharpening methods. \textbf{(b)} The proposed DMAConv adaptively assigns pixels to different branches via a content-adaptive mask, enabling specialized feature modeling. \textbf{(c)} Within each branch, DMAConv generates adaptive convolution kernels based on pixel statistics, dynamically adjusting receptive fields and weight distributions to capture regional heterogeneity better.
}
    \label{fig:toutu}
    %\vspace{-8pt}
\end{figure}

\vspace{-18pt}
\section{Introduction}
\label{sec:intro}
High-resolution multispectral (HRMS) images are indispensable in fields such as urban planning, precision agriculture, and environmental monitoring\cite{xu2022coco,zhuo2022deep}. However, physical sensor constraints make it challenging to acquire images with simultaneously high spatial and spectral resolution. Modern satellites (\eg, WorldView) work around this by capturing two complementary images: low-resolution multispectral (LRMS) and high-resolution panchromatic (PAN). Pansharpening technology has emerged to address this challenge by fusing these two images to generate an HRMS image with both high spatial resolution and high spectral fidelity. As a fundamental remote sensing task, the quality of pansharpening has a direct impact on downstream applications.

Over the past few decades, pansharpening methods have undergone a significant evolution, transitioning from traditional approaches to modern deep learning-based paradigms. Traditional techniques are broadly classified into three main categories: Component Substitution (CS)\cite{choi2010new,vivone2019robust}, Multi-Resolution Analysis (MRA)\cite{vivone2013contrast,vivone2018full}, and Variational Optimization (VO)\cite{fu2019variational,tian2021variational}. In recent years, deep learning (DL-based) methods\cite{xia2025swift,xin2025training} now dominate pansharpening. CNN-based models, from the pioneering PanNet\cite{yangPanNetDeepNetwork2017} to FusionNet\cite{FusionNet}, established a strong foundation by balancing efficiency and performance, but are constrained by local receptive fields. Building on this, Transformer-based models (\eg, CMT\cite{shu2024cmt}) and emerging architectures, such as diffusion models (\eg, SSDiff\cite{zhong2024ssdiff}) have been introduced to capture global dependencies. However, these paradigms introduce significantly higher computational costs and model complexity, and often require larger datasets or face challenges with cross-sensor generalization. As the task is frequently deployed on edge-side platforms, pansharpening inherently requires low computational cost. Therefore, models with overly complex architectures and high computational requirements have limited practical value in our field.

In addition, these methods\cite{huang2025general,he2016deep} have universally overlooked the vast spatial feature heterogeneity inherent in remote sensing imagery: both sprawling natural landscapes and dense urban environments contain large feature-redundant regions (\eg, water bodies, walls) and feature-complex regions (\eg, object edges, coastline). This inherent property creates a profound contradiction with the ``spatially homogeneous" assumption underlying the computational logic of these paradigms. All DL methods are inclined to process all pixels or patches uniformly. This fixed computational strategy leads to an irreconcilable dilemma: in feature-redundant regions, the model imposes unnecessary computational complexity, resulting in immense computational waste, while in feature-complex regions, the fixed operator may lack sufficient expressive power, leading to sub-optimal results. This fundamental conflict between the computational paradigm and the image's properties forces all models built on such homogeneous frameworks into a difficult trade-off between representation power and computational efficiency. Therefore, we pose a core question: \textbf{how to design an architecture that can adapt to the image's spatial heterogeneity while remaining computationally efficient?}

\begin{wrapfigure}{r}{0.5\textwidth} % 这是核心命令
    \centering
    \vspace{-12pt} % 可选：向上微调图片，减少图片上方的空白
    \includegraphics[width=1.0\linewidth]{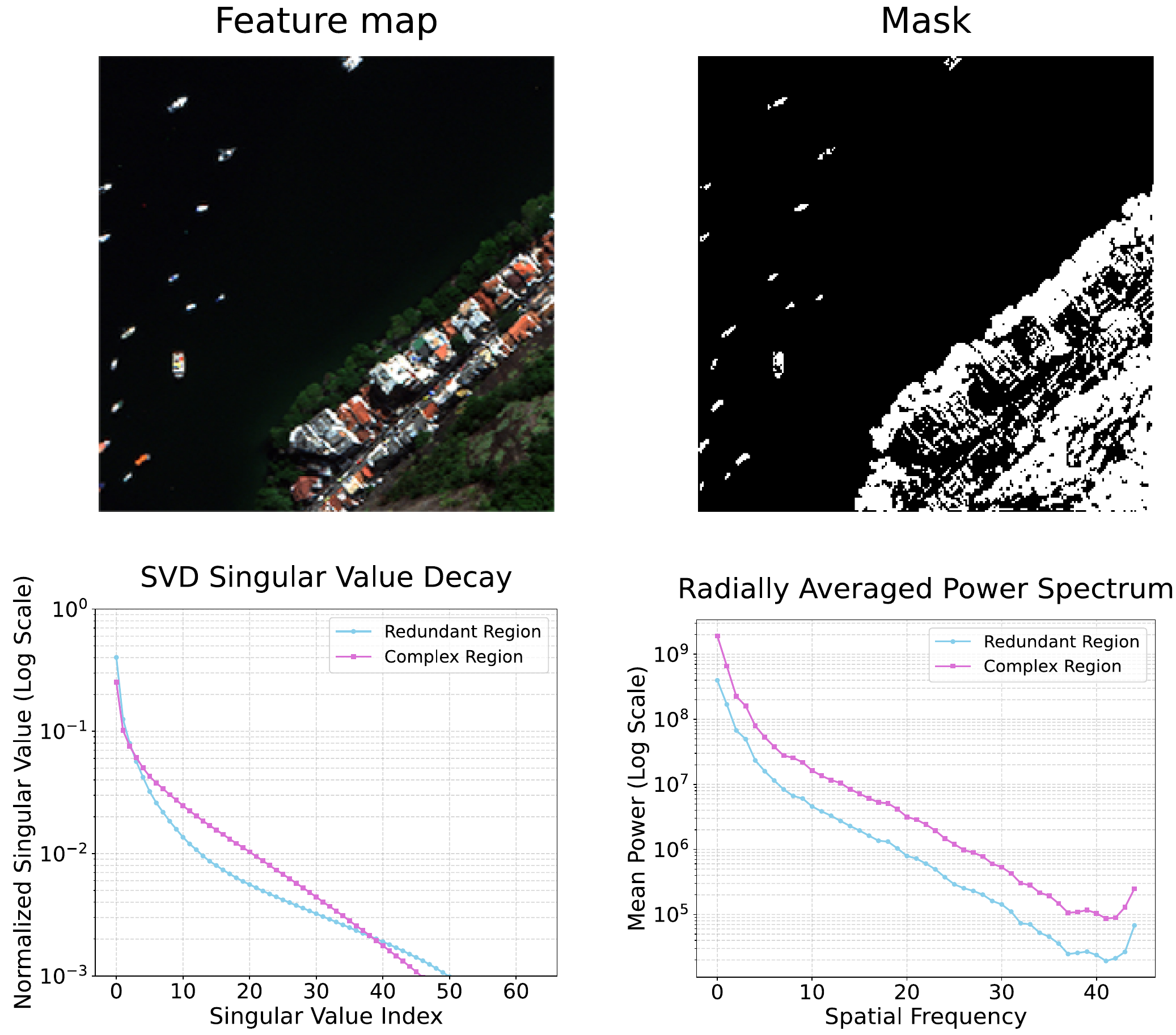}
    \caption{We analyze image patches from typical redundant (\eg, rooftops, water) and complex (\eg, edges, textures) regions. SVD and Fourier analyses validating our core motivation. Redundant regions are shown to be low-rank and dominated by low frequencies, while complex regions are the opposite, thereby motivating our dual-path DMAConv architecture.}
    \label{fig:mmd}
    \vspace{-12pt} % 可选：向下微调图片，减少标题下方的空白
\end{wrapfigure}

Adaptive convolution offers a highly promising avenue for addressing this challenge. By enabling the kernel to be dependent on input content, it can apply specialized processing logic to different regions of the content. However, existing paradigms, while attempting this, expose new challenges and fundamental design flaws. Based on their data processing methods, we can divide them into two types: \textbf{local and non-local strategies}. The Local content-adaptive convolution adapts by modulating kernels based on the regional context of each pixel. Early works (\eg, DFN\cite{jia2016dynamic}) focused on generating a unique, pixel-wise kernel via a sub-network for each pixel; however, this approach introduced significant computational and parameter burdens. In the pansharpening domain, LAGConv\cite{jin2022lagconv} inherited this ideology. It utilizes a simple sub-network to generate a $k \times k$ weight vector for each pixel location, which is then reshaped into a kernel. However, since the sub-network operates on all positions, it brings an immense computational burden and is unable to focus on heterogeneous feature regions. ARConv\cite{wang2025adaptive}, while pivoting to adjust the receptive field shape. Still, it only maps local context to the height and width of the sampling window, failing to leverage the rich local information fully.  

The Non-local content-adaptive convolution leverages broader contextual or non-local information. Some methods (\eg, Dynamic Convolution\cite{chen2020dynamic}) utilize Global Average Pooling (GAP)\cite{Lin2013NetworkIN} to generate global modulation weights; however, this can easily lose fine-grained information from feature-complex regions. The representative work for pansharpening, CANConv\cite{CANNet}, attempts to solve this problem by partitioning the image into regions and generating a shared adaptive kernel for each. However, CANConv suffers from critical bottlenecks: it relies on K-Means clustering\cite{macqueen1967some}, which is extremely computationally inefficient. Furthermore, after partitioning, all clusters still modulate a shared base kernel, which makes it difficult for the model to learn truly isolated and optimal convolutional modes for semantically distinct regions. 
In summary, these paradigms exhibit fundamental design flaws, \textbf{we lack a method that can simultaneously maintain a computationally efficient architecture and effectively address the core problem of spatial heterogeneity.}

A common defect of these methods is their inability to establish a transparent, efficient, and isolated mechanism for handling regions with distinct properties. To this end, we propose a novel adaptive convolution architecture (DMAConv), built upon a paradigm of differentiated computational routing. Our approach utilizes a learned mask to partition image regions, equipping these distinct areas with independent, optimized convolutional modes to prevent mutual interference. This mask-guided, dual-path design is tailored to the bimodal characteristics of the data, applying concise, global processing to feature-redundant areas while allocating intensive, fine-grained computation to feature-complex areas. This asymmetric allocation of computational resources allows our model to achieve significant performance gains while maintaining a total computational load comparable to, or even less than, that of existing methods.

To sum up, our contributions are as follows:
\begin{enumerate}
    \item We propose \textbf{DMAConv}, a novel adaptive convolutional approach that applies distinct, specialized processing modes to feature-redundant and feature-complex regions in remote sensing imagery.

    \item We introduce a new convolutional paradigm for computational triage that partitions features using our mask. This approach accumulates \textbf{computational savings} on redundant regions and reinvests them into complex areas, achieving intelligent allocation of model capacity.

    \item Extensive experiments on multiple benchmark datasets demonstrate that our proposed method achieves superior performance, establishing a new \textbf{state-of-the-art (SOTA)} results on comprehensive spectral fidelity benchmarks.
\end{enumerate}

\begin{figure}[ht]
  \centering
  \includegraphics[width=1\textwidth]{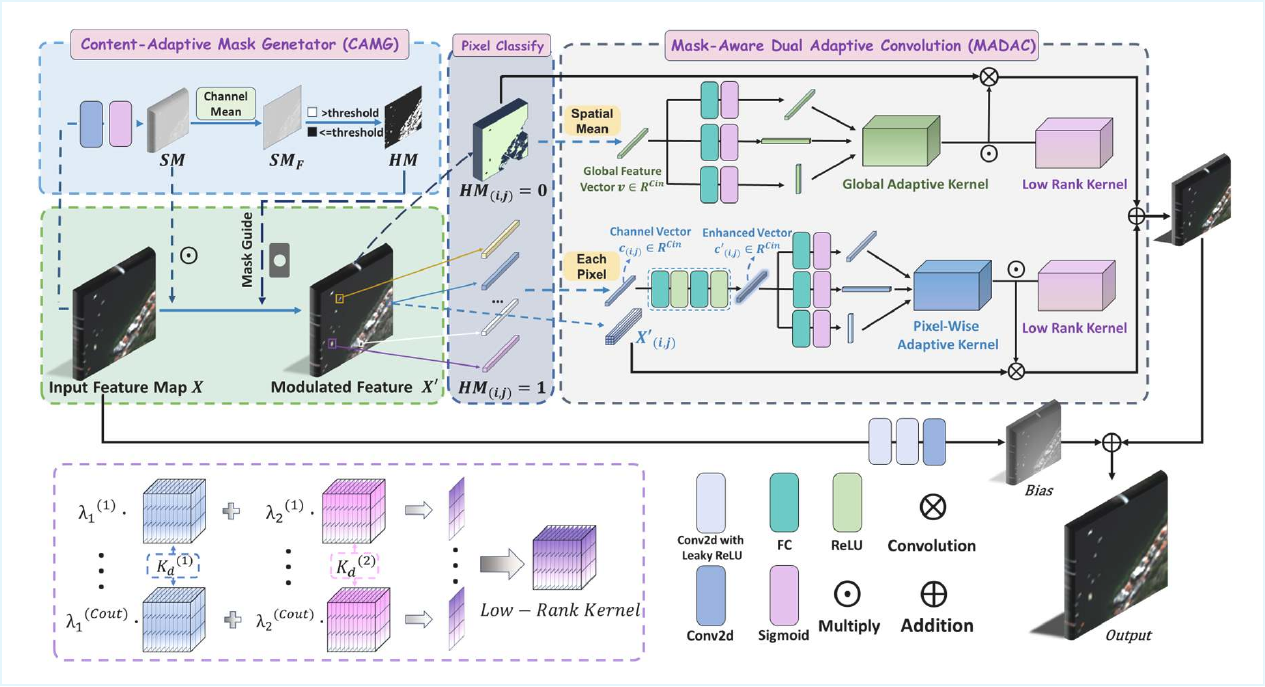}
  \caption{Overview of the proposed DMAConv model. The model consists of two key components: a Content-Adaptive Mask Generator (CAMG) and a Mask-Aware Dual Adaptive Convolution (MADAC). The CAMG first produces hard masks, which guide each pixel into MADAC for adaptive processing. The figure illustrates the overall data flow, as well as the generation of masks, adaptive kernels, and low-rank kernels.
}
  \label{fig:overall_framework}
  %\vspace{-11pt}
\end{figure}
\section{Methods}
\label{sec:method}
\subsection{Motivation}
As previously discussed, in the pansharpening field, images captured by remote sensing satellites possess a distinct characteristic: the majority of regions are feature-redundant and homogeneous. In contrast, a minority of areas are feature-heterogeneous and complex. Traditional convolution, which applies a single convolutional mode across the entire image, cannot optimally process these input features. Existing adaptive convolution paradigms also suffer from design flaws. To leverage information from both feature-redundant and feature-complex regions, rationally allocate computational resources, and process different areas with different convolutional modes, we designed a new adaptive convolution, \textbf{DMAConv}. It first generates a learnable spatial-channel soft mask and a spatial hard mask based on the contextual information of each pixel location, which are used to modulate the adaptive convolution layer's attention to different spaces and channels, and to guide different regions into different processing branches. For feature-redundant areas, we acquire the region's global information and use it to generate a global adaptive kernel. For feature-complex areas, we leverage the contextual information of each location to generate a pixel-wise adaptive kernel. The proposed \textbf{asymmetric} dual-branch architecture addresses the inherent limitations of existing adaptive convolution frameworks and Mixture-of-Experts (MoE)-based approaches, such as CondConv\cite{Yang2019CondConvCP}, by providing a more effective and computationally efficient design tailored to the unique characteristics of remote sensing imagery. In the following section, we will elaborate on the detailed implementation of the module of DMAConv (\cref{fig:overall_framework}).
% \textit{This section introduces our proposed $Bi^2MAC$ module and the network architecture built upon it. The $Bi^2MAC$ module comprises two synergistic submodules: the Content-Adaptive Mask Generator (CAMG) and the Mask-Aware Bimodal Convolution (MABiC). Building upon this module, $Bi^2Net$ is constructed by replacing standard convolutional layers in a baseline model with our $Bi^2MAC$ blocks. This design enables the network to apply differentiated processing strategies to distinct regions, thereby achieving both fine-grained local modeling and global context-aware perception simultaneously.}
\subsection{Dual Mask-Adaptive Convolution}
\subsubsection{Content-Adaptive Mask Generator (CAMG)}
To enable the model to distinguish between feature-redundant and feature-heterogeneous regions, we first design the CAMG. Let the input feature be denoted as $X \in \mathbb{R}^{C_{in} \times H \times W}$, where $C_{in}$ is the number of input channels, $H$ and $W$ are the height and width of the feature map. We aim to generate two outputs: a spatial-channel soft mask $SM \in \mathbb{R}^{C_{in} \times H \times W}$ to preliminarily modulate the model's attention to different spatial and channel-wise information, and a flat, binary hard mask $HM \in \mathbb{R}^{1 \times H \times W}$ to differentiate between the two regions, which guides the subsequent dual-path feature processing.

First, $SM$ is generated by applying a simple convolutional layer followed by a Sigmoid function, $\sigma(x) = \frac{1}{1+e^{-x}}$, which normalizes the feature map to the range $(0, 1)$. This process can be expressed as:
\begin{equation}
  SM = \sigma(W*X + b)
  \label{eq:1}
\end{equation}
where $W$ and $b$ represent the weights and bias of the convolutional layer. Subsequently, $SM$ is element-wise multiplied with the input $X$ to perform initial feature modulation:
\begin{equation}
  X' = X \odot SM
  \label{eq:2}
\end{equation}
Next, to obtain the hard mask, $SM$ is averaged along the channel dimension to get $SM_F \in \mathbb{R}^{1 \times H \times W}$, compressing the channel-wise mask information. We then compute the spatial mean ($\mu$) and standard deviation ($\sigma_s$) of $SM_F$ and define a dynamic threshold $T$:
\begin{equation}
  T = \mu + \alpha \cdot \sigma_s
  \label{eq:important}
\end{equation}
And the binary mask $HM$ is then obtained by binarizing $SM_F$ using this threshold:
\begin{equation}
  HM_{(i, j)} = \begin{cases} 1, & \text{if } SM_{F(i, j)} > T \\ 0, & \text{if } SM_{F(i, j)} \le T \end{cases}
  \label{eq:important}
\end{equation}
where $(i, j)$ denotes the spatial position of the target pixel.

As remote sensing images are composed mainly of feature-redundant areas, with only a minority of regions being feature-heterogeneous, the distribution of $SM_F$ is typically skewed towards low mask values. Therefore, the dynamic threshold strategy, based on mean and standard deviation, can isolate the salient regions with heterogeneous features. The resulting $HM$ acts as a spatial index, clearly classifying each spatial location of the input features. Subsequently, in the MADAC module, the model, guided by $HM$, routes the pixels corresponding to different mask values into separate branches for feature processing.
\subsubsection{Mask-Aware Dual Adaptive Convolution (MADAC)}
This module is designed with two independent sub-networks to process feature-redundant and feature-heterogeneous regions separately. Pixels identified as redundant are assigned to the Compact Branch, while heterogeneous pixels are routed to the Focused Branch.

%\vspace{-12pt}
\paragraph{Compact Branch: Global Adaptive Modeling for Redundant Regions}
For feature-redundant regions (i.e., positions where $HM_{(i,j)}=0$), the corresponding pixels are input to the Compact Branch. This branch first performs global average pooling on these pixels to obtain a global feature vector $v \in \mathbb{R}^{C_{in}}$, which captures the global context. Subsequently, a three-head MLP is used to generate three sets of modulation parameters. This process can be expressed as:
\begin{equation}
 w_{t} = \sigma(f_{t}(v))
  \label{eq:important}
\end{equation}
where $\sigma$ denotes the Sigmoid function. The subscript $t \in \{ci, co, k \times k\}$ ($w_{ci} \in \mathbb{R}^{C_{in}}$, $w_{co} \in \mathbb{R}^{C_{out}}$, $w_{k \times k} \in \mathbb{R}^{k \times k}$) denotes the parameter type, and each $f_{t}$ is an independent fully-connected layer. These parameters then modulate a low-rank convolutional kernel $W_0$ (detailed in the next section) to yield a global adaptive convolutional kernel for redundant regions:
\begin{equation}
 W_0' = W_0 \odot (w_{ci} \otimes w_{co} \otimes w_{k \times k})
  \label{eq:important}
\end{equation}
where $\odot$ denotes element-wise multiplication and $\otimes$ signifies dimension expansion and matching operations. Finally, the Compact Branch processes the features from the redundant regions using this single adaptive kernel.

%\vspace{-12pt}
\paragraph{Focused Branch: Fine-Grained Adaptive Modeling for Heterogeneous Regions}
For feature-heterogeneous regions (i.e., positions where $HM_{(i,j)}=1$), the corresponding pixels are sent to the Focused Branch for pixel-wise adaptive modeling.

First, the channel vector $c_{(i,j)}$ at each spatial position $(i,j)$ is mapped through two fully-connected layers to obtain a feature-enhanced vector:
\begin{equation}
c'_{(i,j)} = \text{ReLU}(\text{FC}_2(\text{ReLU}(\text{FC}_1(c_{(i,j)}))))
  \label{eq:important}
\end{equation}
A three-head MLP then generates three sets of local context-adaptive 
weights for each position:
\begin{equation}
  w_{t(i,j)} = \sigma(f'_{t}(c'_{(i,j)}))
  \label{eq:local_adapter}
\end{equation}
where the subscript \textbf{$t \in \{ci, co, k \times k\}$} indicates the specific local weight, and \textbf{$f'_{t}$} represents one of the three independent MLP heads (distinct from the $f_{t}$ heads). These weights modulate a separate low-rank kernel $W_1$, generating an independent adaptive convolutional kernel $W'_{1(i,j)}$ for each specific position:
\begin{equation}
W'_{1(i,j)} = W_1 \odot (w_{ci(i,j)} \otimes w_{co(i,j)} \otimes w_{k \times k(i,j)})
  \label{eq:important}
\end{equation}
Subsequently, each pixel is processed by its corresponding adaptive kernel. Finally, the outputs $O \in \mathbb{R}^{C_{out} \times H \times W}$ from both branches are reassembled based on the hard mask $HM$. This complete bimodal convolution process, which applies the globally-adaptive kernel $W'_0$ to redundant regions ($HM=0$) and the pixel-wise adaptive kernel $W'_{1(i,j)}$ to the heterogeneous areas ($HM=1$), is formally defined as:
\begin{equation}
O = \begin{cases}
  X' * W'_1 + b_1, & \text{if } HM = 1 \\
  X' * W'_0 + b_0, & \text{if } HM = 0
\end{cases}
  \label{eq:important}
\end{equation}
where $W'_1$ represents the pixel-wise adaptive kernels applied to their respective regions, the symbol $*$ denotes the convolution operation. $b_0$ and $b_1$ are the respective biases for each branch's low-rank kernel. Furthermore, to alleviate boundary artifacts caused by the hard partitioning, DMAConv generates a global bias term by processing the original input feature $X$, enabling feature smoothing and consistency enhancement across regions. This bias term $B \in \mathbb{R}^{C_{out} \times H \times W}$ is generated by a lightweight convolutional block $f_{bias}$ (three $3 \times 3$ convolutions) applied directly to the original input $X$:
\begin{equation}
B = f_{bias}(X)
  \label{eq:important}
\end{equation}
The final output of the module is:
\begin{equation}
O_{(i,j)} + B_{(i,j)}
  \label{eq:important}
\end{equation}
\subsubsection{Independent Low-Rank Kernel}
To enable the two branches to learn complementary feature processing modes, each branch is equipped with a set of weight-independent low-rank convolutional kernels\cite{chen2022spanconv}, in addition to their respective adaptive modulation parameters. The core idea of this low-rank kernel is to significantly reduce the number of kernel parameters while maintaining sufficient expressive power through the synergistic learning of Depth-wise Navigators and Combination Coefficients.

Specifically, the low-rank kernel is composed of two sets of depth-wise navigators $K_d \in \mathbb{R}^{1 \times C_{in} \times k \times k}$ and two sets of combination coefficients $\lambda \in \mathbb{R}^{C_{out} \times C_{in} \times 1 \times 1}$. Each low-rank component maps from the coefficients $\lambda$ to the full kernel via a linear combination of the navigators $K_d$
\begin{equation}
W^{(n)} = \lambda^{(n)} \otimes K_d^{(n)}, \quad W^{(n)} \in \mathbb{R}^{C_{out} \times C_{in} \times k \times k}
  \label{eq:important}
\end{equation}
The final complete low-rank kernel is obtained by summing the two generated kernel components:
\begin{equation}
W_0 = W^{(1)} + W^{(2)}
  \label{eq:important}
\end{equation}
This structural design significantly reduces parameter count while preserving ample representational flexibility. Since the Compact Branch processes redundant inputs and the Focused Branch handles sparse heterogeneous ones, equipping each with an independent low-rank kernel ensures sufficient expressive capacity, drastically reducing parameters and avoiding overfitting.

\subsection{DMANet Architecture}
We construct our DMANet by replacing the standard convolutional layers within the ResBlocks of a renowned U-Net\cite{ronneberger2015u} backbone with our designed DMAConv modules. We selected the U-Net architecture because it is a widely used baseline and highly effective at fusing multi-scale information for image reconstruction. Since the specific U-Net implementation is not our primary innovation, a detailed architectural diagram is provided in the supplementary material.

\section{Experiment}
\label{experi}
\subsection{Datasets, Metrics, and Training Details}

\paragraph{Datasets}
Our experiments are conducted using three widely used high-resolution remote sensing datasets: WorldView-3 (WV3), QuickBird (QB), and GaoFen-2 (GF2). All experiments adhere to the established Wald’s protocol\cite{wald1997fusion}, which is the recognized standard in the remote sensing fusion field. The datasets and associated data processing methods are sourced from the PanCollection\cite{deng2022machine}.

\begin{table}[t]
  \centering 
  \caption{Mean and standard deviation values of the benchmark results on 20 reduced-resolution samples from the QB and GF2 datasets. Best results: \textbf{bold}, second-best: \underline{underline}.} 
  \label{tab:qb_gf2_reduced_results}
  \vspace{-10pt}
  % 使用 \resizebox 将整个 tabular 环境缩放到与文本宽度一致
  % 注意：resizebox 会同时缩放字体大小以适应宽度
  \resizebox{\linewidth}{!}{%
    \begin{tabular}{
      >{\centering\arraybackslash}p{2.2cm}|
      >{\centering\arraybackslash}p{2.1cm} 
      >{\centering\arraybackslash}p{2.1cm} 
      >{\centering\arraybackslash}p{2.1cm}|
      >{\centering\arraybackslash}p{2.1cm}
      >{\centering\arraybackslash}p{2.1cm}
      >{\centering\arraybackslash}p{2.1cm}
   }
      \toprule
      \multirow{2}{*}{Method} & \multicolumn{3}{c|}{QB Reduced: Avg$\pm$std} & \multicolumn{3}{c}{GF2 Reduced: Avg$\pm$std} \\
      \cmidrule{2-4}\cmidrule{5-7}
       & SAM$\downarrow$ & ERGAS$\downarrow$ & Q4$\uparrow$
       & SAM$\downarrow$ & ERGAS$\downarrow$ & Q4$\uparrow$ \\
      \midrule
          BT-H & 7.403$\pm$1.618 & 7.620$\pm$0.864 & 0.819$\pm$0.091  & 1.726$\pm$0.368 & 1.582$\pm$0.403 & 0.910$\pm$0.027  \\
         
          C-BDSD & 8.113$\pm$1.985 & 7.826$\pm$0.863 & 0.826$\pm$0.096  & 1.915$\pm$0.355 & 1.869$\pm$0.449 & 0.901$\pm$0.026  \\
       
          BDSD-PC & 8.195$\pm$2.013 & 7.575$\pm$0.803 & 0.827$\pm$0.090  & 1.760$\pm$0.371 & 1.719$\pm$0.430 & 0.884$\pm$0.035  \\
          
          MTF-GLP & 8.007$\pm$1.835 & 7.648$\pm$0.758 & 0.826$\pm$0.083  & 1.772$\pm$0.410 & 1.794$\pm$0.439 & 0.875$\pm$0.042 \\
      
          MTF-GLP-FS & 7.978$\pm$1.869 & 7.744$\pm$0.806 & 0.820$\pm$0.086  & 1.757$\pm$0.397 & 1.685$\pm$0.376 & 0.889$\pm$0.034  \\
        
          MF & 8.109$\pm$1.928 & 8.361$\pm$1.055 & 0.799$\pm$0.081 & 1.754$\pm$0.381 & 1.832$\pm$0.366 & 0.880$\pm$0.033  \\
          \midrule
          PanNet & 5.767$\pm$1.178 & 5.859$\pm$0.888 & 0.885$\pm$0.092  & 0.997$\pm$0.212 & 0.919$\pm$0.191 & 0.967$\pm$0.010  \\
      
          FusionNet & 4.851$\pm$0.889 & 4.125$\pm$0.312 & 0.926$\pm$0.090  & 0.974$\pm$0.212 & 0.908$\pm$0.152 & 0.969$\pm$0.009  \\
     
          LAGConv & 4.543$\pm$0.820 & 3.759$\pm$0.369 & 0.934$\pm$0.089  & 0.839$\pm$0.157 & 0.779$\pm$0.134 & 0.976$\pm$0.009  \\
          
          CANNet & 4.497$\pm$0.834 & 3.694$\pm$0.338 & 0.937$\pm$0.083  & 0.707$\pm$0.148 & 0.630$\pm$0.128 & 0.983$\pm$0.006  \\
  
          ADWM & 4.450$\pm$0.809 & 3.705$\pm$0.346 & 0.937$\pm$0.085  & \underline{0.672$\pm$0.130} & \underline{0.597$\pm$0.107} & \underline{0.985$\pm$0.006}  \\
  
          ARNet & \underline{4.414$\pm$0.806} & \underline{3.632$\pm$0.327} & \underline{0.939$\pm$0.081}  & 0.698$\pm$0.149 & 0.626$\pm$0.128 & 0.983$\pm$0.007  \\
          \midrule % 替换 \hline 为 \midrule
          \textbf{Proposed} & \textbf{4.369}$\pm$\textbf{0.789} & \textbf{3.562}$\pm$\textbf{0.314} & \textbf{0.939}$\pm$\textbf{0.084}  & \textbf{0.645}$\pm$\textbf{0.134} & \textbf{0.561}$\pm$\textbf{0.107} & \textbf{0.986}$\pm$\textbf{0.006}  \\
      \bottomrule
    \end{tabular}%
  }
\vspace{-8pt}
\end{table}

\begin{table}[t]
  \centering 
  \caption{Mean and standard deviation values of the benchmark results on 20 reduced-resolution samples and 20 full-resolution samples from the WV3 dataset. Best results: \textbf{bold}, second-best: \underline{underline}.}
  \label{tab:wv3}
  \vspace{-10pt}
  % 使用 \resizebox 将整个 tabular 环境缩放到与文本宽度一致
  \resizebox{\linewidth}{!}{%
    % 内部的 \small 和 \setlength 命令会先生效，然后 resizebox 再对结果进行整体缩放
    \begin{tabular}{
      >{\centering\arraybackslash}p{2.2cm}|
      >{\centering\arraybackslash}p{2.1cm} 
      >{\centering\arraybackslash}p{2.1cm} 
      >{\centering\arraybackslash}p{2.1cm}|
      >{\centering\arraybackslash}p{2.1cm}
      >{\centering\arraybackslash}p{2.1cm}
      >{\centering\arraybackslash}p{2.1cm}
   }
      \toprule
      \multirow{2}{*}{Method} & \multicolumn{3}{c|}{WV3 Reduced: Avg$\pm$std} & \multicolumn{3}{c}{WV3 Full: Avg$\pm$std} \\
      \cmidrule{2-4}\cmidrule{5-7}
       & SAM$\downarrow$ & ERGAS$\downarrow$ & Q8$\uparrow$
       & $D_\lambda\downarrow$ & $D_s\downarrow$ & HQNR$\uparrow$\\
      \midrule
          BT-H & 4.912$\pm$1.356 & 4.481$\pm$1.445 & 0.821$\pm$0.095  & 0.066$\pm$0.026 & 0.074$\pm$0.038 & 0.866$\pm$0.057  \\
         
          C-BDSD & 6.192$\pm$1.752 & 5.476$\pm$2.440 & 0.799$\pm$0.114  & 0.090$\pm$0.025 & 0.065$\pm$0.016 & 0.851$\pm$0.027  \\
       
          BDSD-PC & 5.568$\pm$1.859 & 4.740$\pm$1.619 & 0.828$\pm$0.096  & 0.063$\pm$0.025 & 0.073$\pm$0.036 & 0.870$\pm$0.054  \\
          
          MTF-GLP & 5.439$\pm$1.721 & 4.852$\pm$1.614 & 0.827$\pm$0.092  & 0.037$\pm$0.013 & 0.065$\pm$0.037 & 0.900$\pm$0.045 \\
      
          MTF-GLP-FS & 5.509$\pm$1.789 & 4.968$\pm$1.610 & 0.821$\pm$0.091  & 0.036$\pm$0.011 & 0.057$\pm$0.029 & 0.910$\pm$0.036  \\
        
          MF & 5.437$\pm$1.689 & 5.171$\pm$1.501 & 0.808$\pm$0.090 & 0.045$\pm$0.012 & 0.056$\pm$0.027 & 0.901$\pm$0.036  \\
          \midrule
          PanNet & 3.613$\pm$0.766 & 2.664$\pm$0.688 & 0.891$\pm$0.093  & 0.017$\pm$0.007 & 0.047$\pm$0.021 & 0.937$\pm$0.027  \\
      
          FusionNet & 3.358$\pm$0.697 & 2.502$\pm$0.665 & 0.905$\pm$0.089  & 0.024$\pm$0.008 & 0.035$\pm$0.015 & 0.943$\pm$0.020  \\
     
          LAGConv & 3.202$\pm$0.661 & 2.341$\pm$0.606 & 0.907$\pm$0.091  & 0.027$\pm$0.011 & 0.037$\pm$0.013 & 0.937$\pm$0.020  \\
          
          CANNet & 2.941$\pm$0.590 & 2.175$\pm$0.532 & 0.920$\pm$0.084  & 0.020$\pm$0.008 & 0.030$\pm$0.008 & \underline{0.951$\pm$0.013}  \\
          
          ADWM & 2.914$\pm$0.589 & 2.145$\pm$0.531 & 0.919$\pm$0.086  & 0.026$\pm$0.011 & \underline{0.029$\pm$0.014} & 0.946$\pm$0.019  \\
  
          ARNet & \underline{2.883$\pm$0.590} & \underline{2.138$\pm$0.528} & \underline{0.921$\pm$0.083}  & \textbf{0.015$\pm$0.006} & \textbf{0.028$\pm$0.007} & \textbf{0.958$\pm$0.010}  \\
          \midrule % 替换 \hline 为 \midrule
          \textbf{Proposed} & \textbf{2.857$\pm$0.576} & \textbf{2.099$\pm$0.522} & \textbf{0.922$\pm$0.085}  & \underline{0.017$\pm$0.006} & 0.034$\pm$0.003 & 0.950$\pm$0.007  \\
      \bottomrule
    \end{tabular}%
  }
\end{table}

\paragraph{Metrics}
To validate the effectiveness of our method, we adopt multiple evaluation metrics. For reduced-resolution assessment, we use the Spectral Angle Mapper (SAM)\cite{vivone2014critical}, Erreur Relative Globale Adimensionnelle de Synthèse (ERGAS)\cite{wald2002data}, and the $Q_{N}$ index\cite{garzelli2009hypercomplex}. For full-resolution assessment, we employ the Hybrid Quality with No Reference (HQNR) indices \cite{arienzo2022full}, specifically $D_{\lambda}$ and $D_s$. Notably, HQNR is derived from $D_{\lambda}$ and $D_s$, providing an assessment of overall fusion quality.

\paragraph{Training Details}
All experiments were conducted on an NVIDIA RTX 3090 GPU. We selected this high-performance, consumer-grade GPU to realistically simulate resource-constrained deployment scenarios, thereby validating the potential of our method for edge-computing applications. All models are trained using the $l_1$ loss, optimized with Adam optimizer\cite{Kingma2014AdamAM}. The batch size is set to 32, and $\alpha$ for threshold computation is fixed at 2. We adopt the Straight-Through Estimator (STE)\cite{Bengio2013EstimatingOP} to approximate gradients through the non-differentiable mask binarization operation. Our implementation is designed to be robust against mode collapse into a single branch. Even if this occurs during training, our method can continue training and escape from such states. The initial learning rate is 0.0006, and it is decayed by a factor of 0.8 every 200 epochs. Training is performed for 600 epochs on the WV3 dataset and 400 epochs on both the QB and GF2 datasets.

\subsection{Comparison with SOTA Methods}
\label{sec:comparison}
We compare our method with various SOTA benchmarks to demonstrate the comprehensiveness of our experiments and the superiority of our approach. Our benchmarks include six traditional methods (BT-H, C-BDSD, BDSD-PC, MTF-GLP, MTF-GLP-FS, MF)\cite{garzelli2008pansharpening, Zhong2017CombiningCS, Vivone2019RobustBS, Aiazzi2006MTFtailoredMA, 8325487, restaino2016fusion} and six deep learning methods. The latter includes classic methods like PanNet, FusionNet, and the recent ADWM. To highlight the advancement of our adaptive convolution, we also compare against the primary adaptive methods in the field: LAGConv, CANNet, and ARNet. The performance of additional baselines, \eg, more CNN-based, Transformer-based, and Diffusion-based methods, is provided in the supplementary materials.

\begin{figure}[h]
    \centering
    \includegraphics[width=1\linewidth]{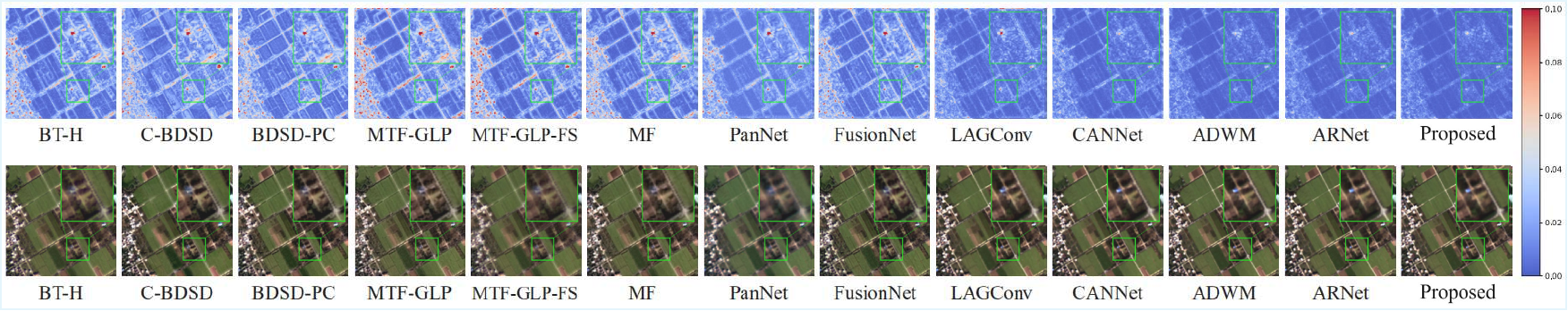}
    \caption{ Visual Fusion image and Error maps on GF2 dataset (reduced data). For the error maps, blue indicates low error.}
    \label{fig:enter-label}
    %\vspace{-8pt}
\end{figure}

\cref{tab:qb_gf2_reduced_results} and \cref{tab:wv3} present the quantitative comparison of our method against these leading-edge techniques on the three datasets. Notably, our method exhibits superior and highly consistent performance across all datasets, outperforming all previous models on the vast majority of key metrics. Specifically, on the challenging 8-band WV3 dataset, our method achieves the optimal score on the critical spectral metric SAM with 2.857, while matching or exceeding the state-of-the-art on other metrics. Our superiority is particularly evident on the 4-band datasets: on the GF2 dataset, our SAM (0.645) and ERGAS (0.561) metrics significantly outperform all prior adaptive convolution methods; on the QB dataset, our method likewise surpasses all three Reduced-Resolution metrics. These results robustly confirm that our model has achieved a new SOTA on the vast majority of critical fidelity metrics. To clearly illustrate the performance of each method, we also provide visual comparisons of fusion results and their corresponding error maps in \cref{fig:enter-label}.

%\vspace{-2pt}
\subsection{Efficiency and Performance Analysis}
\begin{wrapfigure}{r}{0.5\columnwidth}
  \vspace{-20pt} % 向上移动图片，减少图片上方的空白
  \centering
  \includegraphics[width=1.0\linewidth]{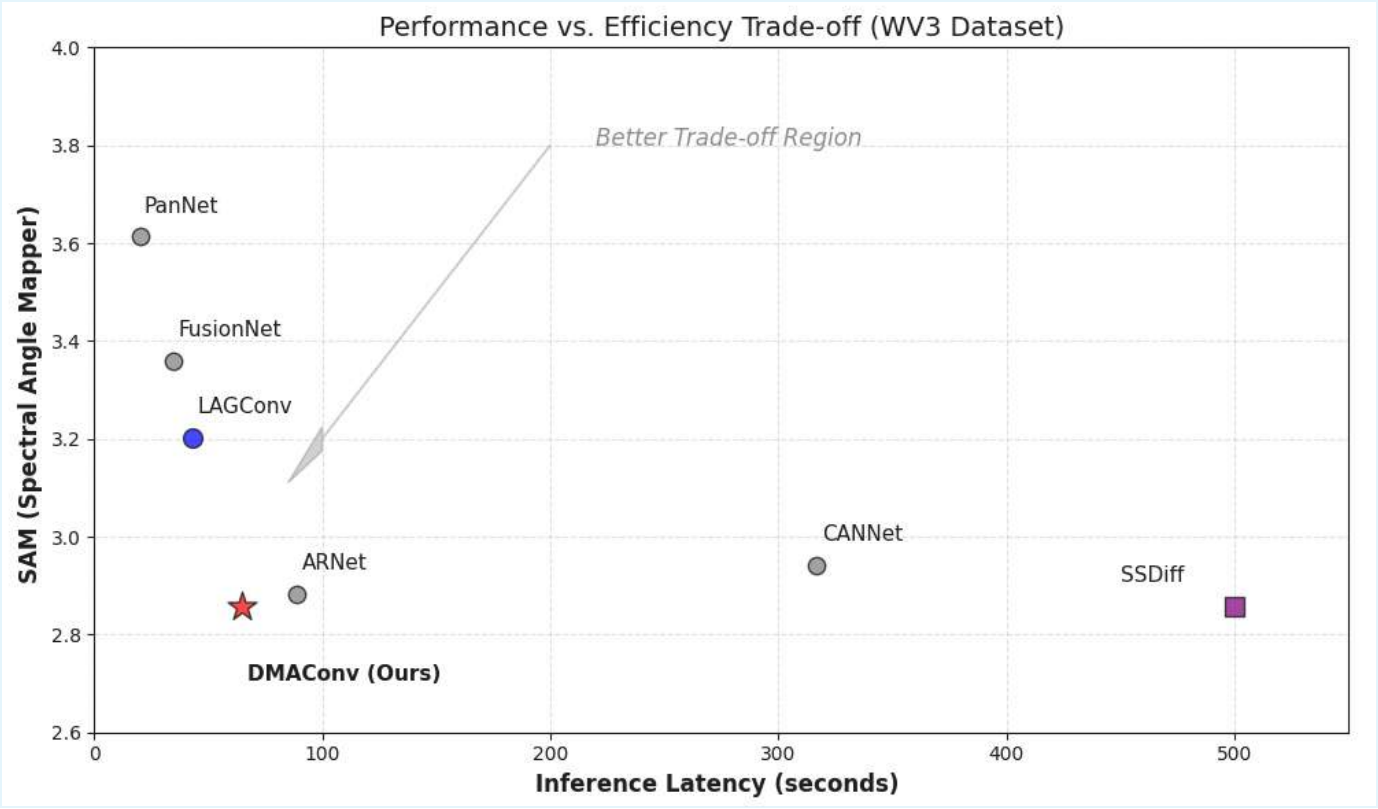}
  \caption{The performance-efficiency Pareto front, our model (Star) achieves the best balance.}
  \label{fig:efficiency}
  \vspace{-16pt} % 向上移动文字，减少图片下方的空白
\end{wrapfigure}

To evaluate the true computational utility of our architecture, we analyze efficiency as a trade-off between inference latency and reconstruction fidelity (\eg, SAM). As illustrated in the Pareto front (\cref{fig:efficiency}), existing methods suffer from extreme imbalances. Legacy lightweight operators like LAGConv, while fast (43s), hit a severe performance ceiling (SAM 3.202) due to their simplistic uniform designs. And recent generative model SSDiff incur unacceptable latency (>500 s).DMAConv successfully breaks this bottleneck. 

The key to our optimal efficiency lies in the asymmetric ``computational triage'' of our module. The CAMG routes approximately 85\% of pixels (feature-redundant) to a low-cost global path, while concentrating resources on only 15\% of heterogeneous pixels (details on this ratio are provided in the supplementary material). This intelligent allocation contrasts sharply with the brute-force redundancy of LAGConv and the iterative K-Means bottleneck of CANNet. 
Consequently, DMANet achieves the best spectral fidelity (SAM 2.857) with only a moderate latency (65s). It remains the fastest among all high-performance methods (outpacing ARNet at 89s and CANNet at 317s), strictly occupying the optimal frontier on the Pareto plot.

\begin{wrapfigure}{r}{0.4\columnwidth}
    \vspace{-4pt} % 调整图片与上方文本的垂直距离
    \centering
    \includegraphics[width=1.0\linewidth]{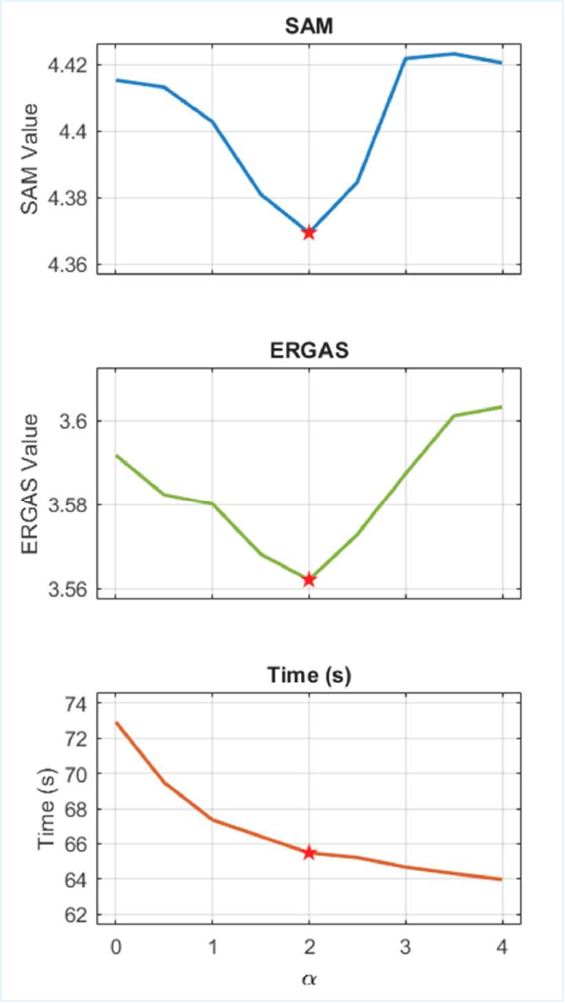}
    \setlength{\abovecaptionskip}{-15pt}
    \caption{Effect of $\alpha$ on SAM, ERGAS, and training time on the QB dataset. The three vertically stacked plots share $\alpha$ as the horizontal axis.$\alpha=2$ (marked with a star) achieves the best performance.}
    \label{fig:alpha}
    \vspace{-44pt} % 调整标题与下方文本的垂直距离
\end{wrapfigure}

\paragraph{Analysis of the Hyperparameter $\alpha$.}
The hyperparameter $\alpha$ plays a critical role in determining the routing threshold of the dual-branch architecture. To analyze its impact on both performance and efficiency, we conduct experiments on the QB dataset while keeping all other settings fixed. Specifically, $\alpha$ is varied from 0 to 3 with an interval of 0.5. Model performance is evaluated using SAM and ERGAS, and computational efficiency is measured by the average per-epoch training time over the first 100 epochs.
As shown in \cref{fig:alpha}, both SAM and ERGAS exhibit a clear U-shaped trend as $\alpha$ increases, first decreasing and then rising, with the minimum achieved at $\alpha = 2$. This indicates that the model attains its best fusion quality at this point. When $\alpha$ is small, the threshold is low, causing a large proportion of pixels—including those from homogeneous regions—to be routed to the focused branch. The redundant information from homogeneous regions interferes with the processing of heterogeneous areas, preventing the focused branch from concentrating on difficult-to-fuse regions. Conversely, when $\alpha$ is large, the threshold becomes overly restrictive, allowing only a small number of pixels to enter the focused branch. As a result, heterogeneous regions are insufficiently exploited, leading to degraded performance.

Regarding computational cost, the training time decreases sharply at first and then declines more gradually. The initial acceleration arises because increasing $\alpha$ reduces the number of pixels processed by the computationally intensive focused branch, thereby lowering the memory throughput required for clustering and aggregation. However, when $\alpha > 2$, the threshold is already sufficiently high such that further increases in $\alpha$ do not significantly reduce the routing ratio. Consequently, the efficiency gain becomes marginal.

Overall, $\alpha = 2$ provides the best balance between fusion performance and computational efficiency. This trend also holds for other datasets, and the corresponding experiments can be found in the supplementary material.

\subsection{Discussion}

\paragraph{Effectiveness of Low-Rank Kernels and Independent Weights}
To evaluate the roles of low-rank kernels and independent base kernels in DMAConv, we conduct two comparisons: (1) replacing the low-rank kernels (LRK) with standard convolutions, and (2) forcing both branches to share the same LRK weights. As shown in \cref{tab:lrk}, substituting LRKs with standard kernels results in apparent performance degradation and overfitting, indicating that redundant pixels in the compact branch and sparse heterogeneous pixels in the focused branch are both sufficiently modeled by the low-rank formulation, whereas standard convolutions introduce unnecessary parameter redundancy. Additionally, shared kernel weights also reduce performance by limiting branch-specific adaptation to regional variations in features. These results confirm that using separate low-rank base kernels enables each branch to capture redundant and heterogeneous regions better, thereby improving fusion quality.

\begin{table}[t]
    % 第一个 minipage：放置第一个表格,占页面宽度 0.48
        \footnotesize  % 设置表格内容字体大小为 small
        \centering  % 表格居中
        \caption{Performance comparison of models without low-rank kernels and with shared base kernel weights, trained and tested under identical hyperparameters. Best: \textbf{bold}, second-best: \underline{underline}.}
        %\vspace{-12pt}
        \label{tab:lrk}
        \setlength{\tabcolsep}{2pt}  % 行间距
        \renewcommand\arraystretch{0.97}  % 行间距系数
        
        \begin{tabular}{
             >{\centering\arraybackslash}p{2.3cm}| % 第1列
             >{\centering\arraybackslash}p{1.9cm} % 第2列
             >{\centering\arraybackslash}p{1.9cm} % 第3列
             >{\centering\arraybackslash}p{1.9cm} % 第4列
        }
            \toprule
            \multirow{2}{*}{Method} & \multicolumn{3}{c}{WV3: Avg$\pm$std}  \\
            \cmidrule{2-4}
             & SAM$\downarrow$ & ERGAS$\downarrow$  & Q8$\uparrow$ \\
            \midrule
                No LRK & 2.874$\pm$0.578 & 2.122$\pm$0.523 & 0.921$\pm$0.084  \\
                \midrule 
                Shared Weights & \underline{2.873$\pm$0.585} & \underline{2.107$\pm$0.526} & \underline{0.921$\pm$0.084}  \\
                \midrule 
                \textbf{Proposed} & \textbf{2.857$\pm$0.576} & \textbf{2.099$\pm$0.522} & \textbf{0.922$\pm$0.085}   \\
            \bottomrule
        \end{tabular}
\end{table}

%\vspace{-8pt}
\paragraph{Ablation Study}
To evaluate the effectiveness of each module, we conducted a series of ablation experiments, with results summarized in \cref{tab:ablation}. For the ``no compact branch" and ``no focused branch" settings, we retained soft-mask generation and modulation, but adjusted the threshold so that all pixels were forced into either the compact or focused branch. As shown in \cref{tab:ablation}, removing either branch causes apparent performance degradation, indicating the complementary and necessary roles of the dual-branch structure.
In the no CAMG experiment, we removed the Content-Adaptive Mask Generator. Since the hard mask derives from the soft mask, CAMG removal prevents the generation of a meaningful hard mask; thus, a random one was used instead, with 15\% of pixels assigned to the focused branch. This leads to a significant performance drop, confirming the importance of CAMG in pixel-wise assignment and feature adaptivity.

%\vspace{-10pt}
\paragraph{Replacing Standard Convolutions}
We design DMAConv as a plug-and-play adaptive convolutional module. To assess its effectiveness and generality as an independent component, we replaced the standard convolution layers in several representative pansharpening CNN architectures with DMAConv. As shown in \cref{tab:plug}, substituting the four ResBlocks in FusionNet with DMAConv-ResBlocks leads to a substantial performance gain. Likewise, replacing the locally adaptive convolution (LAGConv) with DMAConv further improves performance. Moreover, integrating DMAConv in place of ARConv within ARNet also yields consistent enhancements, further demonstrating the superiority of our module. Collectively, these experiments highlight the strong adaptability and robustness of the proposed design.

\begin{table}[t] % 在单栏模板中使用 table 即可
  \centering 
  \caption{Ablation study and replacing standard convolutions on the WV3 dataset.}
  % --- 第一个子表格 (Ablation Study) ---
  \begin{subtable}[t]{0.465\linewidth} % 分配页面宽度的 49%
    \centering
    \caption{Ablation study on WV3. Best: \textbf{bold}, second-best: \underline{underline}.}
    \label{tab:ablation}
    % 在 subtable 内部，用 resizebox 强制缩放 tabular
    \resizebox{\linewidth}{!}{%
      \begin{tabular}{
        >{\centering\arraybackslash}m{1.8cm}|
        >{\centering\arraybackslash}m{1.9cm}
        >{\centering\arraybackslash}m{1.9cm}
        >{\centering\arraybackslash}m{1.9cm}
      }
        \toprule
        \multirow{2}{*}{Ablation} & \multicolumn{3}{c}{WV3: Avg$\pm$std} \\
        \cmidrule{2-4}
        & SAM$\downarrow$ & ERGAS$\downarrow$ & Q8$\uparrow$ \\
        \midrule
        No Focused Branch & 2.886$\pm$0.580 & \underline{2.115$\pm$0.525} & 0.919$\pm$0.087 \\
        \midrule
        No Compact Branch & \underline{2.875$\pm$0.581} & 2.125$\pm$0.515 & \underline{0.921$\pm$0.084} \\
        \midrule
        No CAMG & 2.899$\pm$0.587 & 2.118$\pm$0.523 & 0.920$\pm$0.086 \\
        \midrule
        \textbf{Proposed} & \textbf{2.857$\pm$0.576} & \textbf{2.099$\pm$0.522} & \textbf{0.922$\pm$0.085} \\
        \bottomrule
      \end{tabular}%
    }
  \end{subtable}%
  \hfill % 弹性空白，使两个表格左右对齐
  % --- 第二个子表格 (Plug-in) ---
  \begin{subtable}[t]{0.522\linewidth} % 分配页面宽度的 49%
    \centering
    \caption{Results of plugging DMAConv into different backbones. Best: \textbf{bold}, second-best: \underline{underline}.}
    \label{tab:plug}
    % 同样，在这里也用 resizebox 强制缩放
    \resizebox{\linewidth}{!}{%
      \begin{tabular}{
        >{\centering\arraybackslash}p{2.4cm}|
        >{\centering\arraybackslash}p{1.9cm}
        >{\centering\arraybackslash}p{1.9cm}
        >{\centering\arraybackslash}p{1.9cm}
      }
        \toprule
        \multirow{2}{*}{Method} & \multicolumn{3}{c}{WV3: Avg$\pm$std} \\
        \cmidrule{2-4}
        & SAM$\downarrow$ & ERGAS$\downarrow$ & Q8$\uparrow$ \\
        \midrule
        FusionNet & 3.358$\pm$0.697 & 2.502$\pm$0.665 & 0.905$\pm$0.089 \\
        DMA-FusionNet & 3.056$\pm$0.631 & 2.273$\pm$0.566 & 0.915$\pm$0.086 \\
        \midrule
        LAGNet & 3.202$\pm$0.661 & 2.341$\pm$0.606 & 0.907$\pm$0.091 \\
        DMA-LAGNet & 2.966$\pm$0.598 & 2.206$\pm$0.575 & 0.918$\pm$0.086 \\
        \midrule
        ARNet & \underline{2.883$\pm$0.590} & \underline{2.138$\pm$0.528} & \underline{0.921$\pm$0.083} \\
        DMA-ARNet & \textbf{2.857$\pm$0.576} & \textbf{2.099$\pm$0.522} & \textbf{0.922$\pm$0.085} \\
        \bottomrule
      \end{tabular}%
    }
  \end{subtable}
  
  %\vspace{-14pt} % 全局的垂直间距调整
\end{table}

\begin{figure}[h]
    \centering
    \includegraphics[width=1\linewidth]{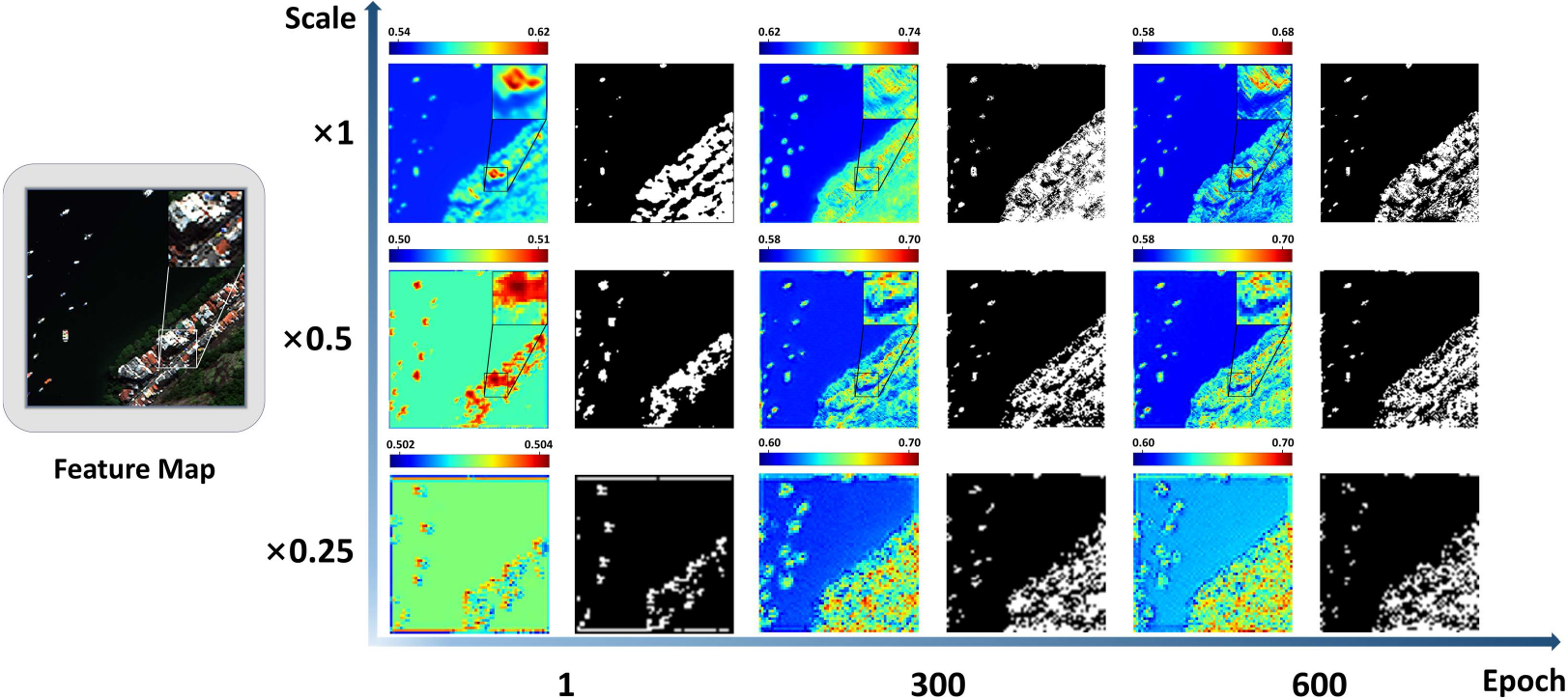}
    \caption{ Visualization of the soft mask flat ($SM_F$) heatmaps and hard mask ($HM$) maps generated by DMAConv at different training stages and network depths. Scale denotes the downsampling ratio in the U-Net. The color intensity in $SM_F$ reflects the model’s attention level to each region, while the black-and-white pattern in $HM$ indicates the branch assignment for pixel-wise processing.}
    \label{fig:maskvis}
\end{figure}

%\vspace{-10pt}
\paragraph{Mask Visualization}
\cref{fig:maskvis} visualizes the distributions of $SM_F$ and $HM$ produced by DMANet across different network depths and training stages. As shown, $SM_F$ consistently exhibits stronger responses in detail-rich and edge-dominant regions, revealing that the model adaptively allocates spatial attention according to local structural complexity. In contrast, the pixel-level $HM$ provides a crisp binary delineation between redundant and heterogeneous regions, confirming its effectiveness in routing pixels into appropriate processing branches. Notably, deeper layers and later training epochs yield progressively sharper, more coherent mask boundaries, indicating that the model learns increasingly reliable region partitioning as optimization proceeds. These observations highlight the strong interpretability of the proposed CAMG module and underscore its central role in enhancing DMANet’s joint structural–spectral modeling capability.

\vspace{-22pt}
\section{Conclusion}
In this paper, we propose DMAConv, a novel adaptive convolutional module that simultaneously enables computational allocation and attention adaptability in a unified manner. The module perceives the spatial distribution of image features and employs a mask-guided mechanism to route pixels into distinct processing paths, allowing heterogeneous and redundant regions to be modeled with appropriate levels of complexity. This design not only enhances regional awareness and feature discrimination, but also maintains high computational efficiency through selective processing and a lightweight low-rank formulation. Extensive experiments across multiple datasets and backbone architectures demonstrate that DMAConv consistently delivers superior performance, confirming its effectiveness, robustness, and plug-and-play versatility as a general adaptive convolutional operator. To ensure the reproducibility of our results and to facilitate future research in the community, we will release the complete source code for our project on GitHub.
\bibliographystyle{splncs04}
\bibliography{main}

@String(CVPR  = {IEEE Conf. Comput. Vis. Pattern Recog.})

@String(ICCV  = {Int. Conf. Comput. Vis.})

@String(AAAI  = {AAAI})

@String(IJCAI = {IJCAI})

@String(CVPR  = {CVPR})

@String(ICCV  = {ICCV})

@inproceedings{yangPanNetDeepNetwork2017,
  title = {{{PanNet}}: {{A Deep Network Architecture}} for {{Pan-Sharpening}}},
  shorttitle = {{{PanNet}}},
  booktitle = {2017 {{IEEE International Conference}} on {{Computer Vision}} ({{ICCV}})},
  author = {Yang, Junfeng and Fu, Xueyang and Hu, Yuwen and Huang, Yue and Ding, Xinghao and Paisley, John},
  year = {2017},
  pages = {1753--1761}
}

@article{FusionNet,
  title={Detail Injection-Based Deep Convolutional Neural Networks for Pansharpening}, 
  journal={IEEE Transactions on Geoscience and Remote Sensing},  
  author={Deng, Liang-Jian and Vivone, Gemine and Jin, Cheng and Chanussot, Jocelyn},  
  year={2021},  
  month={Jul},  
  pages={6995–7010},  
  language={en-US}  }

@article{zhong2024ssdiff,
  title={Ssdiff: Spatial-spectral integrated diffusion model for remote sensing pansharpening},
  author={Zhong, Yu and Wu, Xiao and Cao, Zihan and Dou, Hong-Xia and Deng, Liang-Jian},
  journal={Advances in Neural Information Processing Systems},
  volume={37},
  pages={77962--77986},
  year={2024}
}

@inproceedings{huang2025general,
  title={A General Adaptive Dual-level Weighting Mechanism for Remote Sensing Pansharpening},
  author={Huang, Jie and Chen, Haorui and Ren, Jiaxuan and Peng, Siran and Deng, Liangjian},
  booktitle={Proceedings of the Computer Vision and Pattern Recognition Conference},
  pages={7447--7456},
  year={2025}
}

@inproceedings{jin2022lagconv,
  title={LAGConv: Local-context adaptive convolution kernels with global harmonic bias for pansharpening},
  author={Jin, Zi-Rong and Zhang, Tian-Jing and Jiang, Tai-Xiang and Vivone, Gemine and Deng, Liang-Jian},
  booktitle={Proceedings of the AAAI conference on artificial intelligence},
  volume={36},
  number={1},
  pages={1113--1121},
  year={2022}
}

@inproceedings{wang2025adaptive,
  title={Adaptive Rectangular Convolution for Remote Sensing Pansharpening},
  author={Wang, Xueyang and Zheng, Zhixin and Shao, Jiandong and Duan, Yule and Deng, Liang-Jian},
  booktitle={Proceedings of the Computer Vision and Pattern Recognition Conference},
  pages={17872--17881},
  year={2025}
}

@InProceedings{CANNet,
    author    = {Duan, Yule and Wu, Xiao and Deng, Haoyu and Deng, Liang-Jian},
    title     = {Content-Adaptive Non-Local Convolution for Remote Sensing Pansharpening},
    booktitle = {Proceedings of the IEEE/CVF Conference on Computer Vision and Pattern Recognition (CVPR)},
    month     = {June},
    year      = {2024},
    pages     = {27738-27747}
}

@inproceedings{chen2020dynamic,
  title={Dynamic convolution: Attention over convolution kernels},
  author={Chen, Yinpeng and Dai, Xiyang and Liu, Mengchen and Chen, Dongdong and Yuan, Lu and Liu, Zicheng},
  booktitle={Proceedings of the IEEE/CVF conference on computer vision and pattern recognition},
  pages={11030--11039},
  year={2020}
}

@article{Zhong2017CombiningCS,
  title={Combining Component Substitution and Multiresolution Analysis: A Novel Generalized BDSD Pansharpening Algorithm},
  author={Shengwei Zhong and Ye Zhang and Yushi Chen and Di Wu},
  journal={IEEE Journal of Selected Topics in Applied Earth Observations and Remote Sensing},
  year={2017},
  volume={10},
  pages={2867-2875}
}

@article{garzelli2008pansharpening,
  title={Pansharpening of multispectral images: a critical review and comparison},
  author={Garzelli, Andrea and Alparone, Luciano and Nencini, Filippo and Baronti, Stefano},
  journal={EURASIP Journal on Advances in Signal Processing},
  volume={2008},
  pages={1--13},
  year={2008},
  publisher={Hindawi Publishing Corporation}
}

@inproceedings{Aiazzi2006MTFtailoredMA,
  title={MTF-tailored multi-resolution analysis for fusion of multi-spectral and panchromatic images},
  author={Bruno Aiazzi and Luciano Alparone and Stefano Baronti and Andrea Garzelli and Filippo Nencini and Massimo Selva},
  year={2006}
}

@article{Vivone2019RobustBS,
  title={Robust Band-Dependent Spatial-Detail Approaches for Panchromatic Sharpening},
  author={Gemine Vivone},
  journal={IEEE Transactions on Geoscience and Remote Sensing},
  year={2019},
  volume={57},
  pages={6421-6433}
}

@ARTICLE{8325487,
  author={Vivone, Gemine and Restaino, Rocco and Chanussot, Jocelyn},
  journal={IEEE Transactions on Image Processing}, 
  title={Full Scale Regression-Based Injection Coefficients for Panchromatic Sharpening}, 
  year={2018},
  volume={27},
  number={7},
  pages={3418-3431}
}

@article{restaino2016fusion,
  title={Fusion of multispectral and panchromatic images based on morphological operators},
  author={Restaino, Rocco and Vivone, Gemine and Dalla Mura, Mauro and Chanussot, Jocelyn},
  journal={IEEE Transactions on Image Processing},
  volume={25},
  number={6},
  pages={2882--2895},
  year={2016},
  publisher={IEEE}
}

@article{shu2024cmt,
  title={Cmt: Cross modulation transformer with hybrid loss for pansharpening},
  author={Shu, Wen-Jie and Dou, Hong-Xia and Wen, Rui and Wu, Xiao and Deng, Liang-Jian},
  journal={IEEE Geoscience and Remote Sensing Letters},
  year={2024},
  publisher={IEEE}
}

@article{choi2010new,
  title={A new adaptive component-substitution-based satellite image fusion by using partial replacement},
  author={Choi, Jaewan and Yu, Kiyun and Kim, Yongil},
  journal={IEEE transactions on geoscience and remote sensing},
  volume={49},
  number={1},
  pages={295--309},
  year={2010},
  publisher={IEEE}
}

@article{vivone2019robust,
  title={Robust band-dependent spatial-detail approaches for panchromatic sharpening},
  author={Vivone, Gemine},
  journal={IEEE transactions on Geoscience and Remote Sensing},
  volume={57},
  number={9},
  pages={6421--6433},
  year={2019},
  publisher={IEEE}
}

@article{xu2022coco,
  title={COCO-Net: A dual-supervised network with unified ROI-loss for low-resolution ship detection from optical satellite image sequences},
  author={Xu, Qizhi and Li, Yuan and Zhang, Mingjin and Li, Wei},
  journal={IEEE Transactions on Geoscience and Remote Sensing},
  volume={60},
  pages={1--15},
  year={2022},
  publisher={IEEE}
}

@article{vivone2013contrast,
  title={Contrast and error-based fusion schemes for multispectral image pansharpening},
  author={Vivone, Gemine and Restaino, Rocco and Dalla Mura, Mauro and Licciardi, Giorgio and Chanussot, Jocelyn},
  journal={IEEE Geoscience and Remote Sensing Letters},
  volume={11},
  number={5},
  pages={930--934},
  year={2013},
  publisher={IEEE}
}

@article{vivone2018full,
  title={Full scale regression-based injection coefficients for panchromatic sharpening},
  author={Vivone, Gemine and Restaino, Rocco and Chanussot, Jocelyn},
  journal={IEEE Transactions on Image Processing},
  volume={27},
  number={7},
  pages={3418--3431},
  year={2018},
  publisher={IEEE}
}

@inproceedings{fu2019variational,
  title={A variational pan-sharpening with local gradient constraints},
  author={Fu, Xueyang and Lin, Zihuang and Huang, Yue and Ding, Xinghao},
  booktitle={Proceedings of the IEEE/CVF Conference on Computer Vision and Pattern Recognition},
  pages={10265--10274},
  year={2019}
}

@article{tian2021variational,
  title={Variational pansharpening by exploiting cartoon-texture similarities},
  author={Tian, Xin and Chen, Yuerong and Yang, Changcai and Ma, Jiayi},
  journal={IEEE Transactions on Geoscience and Remote Sensing},
  volume={60},
  pages={1--16},
  year={2021},
  publisher={IEEE}
}

@article{xia2025swift,
  title={SWIFT: A General Sensitive Weight Identification Framework for Fast Sensor-Transfer Pansharpening},
  author={Xia, Zeyu and Sun, Chenxi and Xin, Tianyu and Zeng, Yubo and Chen, Haoyu and Deng, Liang-Jian},
  journal={arXiv preprint arXiv:2507.20311},
  year={2025}
}

@article{xin2025training,
  title={Training and Inference within 1 Second--Tackle Cross-Sensor Degradation of Real-World Pansharpening with Efficient Residual Feature Tailoring},
  author={Xin, Tianyu and Xiao, Jin-Liang and Xia, Zeyu and Yin, Shan and Deng, Liang-Jian},
  journal={arXiv preprint arXiv:2508.07369},
  year={2025}
}

@inproceedings{he2016deep,
  title={Deep residual learning for image recognition},
  author={He, Kaiming and Zhang, Xiangyu and Ren, Shaoqing and Sun, Jian},
  booktitle={Proceedings of the IEEE conference on computer vision and pattern recognition},
  pages={770--778},
  year={2016}
}

@article{jia2016dynamic,
  title={Dynamic filter networks},
  author={Jia, Xu and De Brabandere, Bert and Tuytelaars, Tinne and Gool, Luc V},
  journal={Advances in neural information processing systems},
  volume={29},
  year={2016}
}

@article{Lin2013NetworkIN,
  title={Network In Network},
  author={Min Lin and Qiang Chen and Shuicheng Yan},
  journal={CoRR},
  year={2013},
  volume={abs/1312.4400}
}

@misc{macqueen1967some,
  title={Some methods for classification and analysis of multivariate observations. Volume 1 of Proceedings of the Fifth Berkeley Symposium on Mathematical statistics and probability},
  author={MacQueen, J},
  year={1967},
  publisher={Berkeley}
}

@article{zhuo2022deep,
  title={A deep-shallow fusion network with multidetail extractor and spectral attention for hyperspectral pansharpening},
  author={Zhuo, Yu-Wei and Zhang, Tian-Jing and Hu, Jin-Fan and Dou, Hong-Xia and Huang, Ting-Zhu and Deng, Liang-Jian},
  journal={IEEE Journal of Selected Topics in Applied Earth Observations and Remote Sensing},
  volume={15},
  pages={7539--7555},
  year={2022},
  publisher={IEEE}
}

@inproceedings{ronneberger2015u,
  title={U-net: Convolutional networks for biomedical image segmentation},
  author={Ronneberger, Olaf and Fischer, Philipp and Brox, Thomas},
  booktitle={International Conference on Medical image computing and computer-assisted intervention},
  pages={234--241},
  year={2015},
  organization={Springer}
}

@book{wald2002data,
  title={Data fusion: definitions and architectures: fusion of images of different spatial resolutions},
  author={Wald, Lucien},
  year={2002},
  publisher={Presses des MINES}
}

@article{garzelli2009hypercomplex,
  title={Hypercomplex quality assessment of multi/hyperspectral images},
  author={Garzelli, Andrea and Nencini, Filippo},
  journal={IEEE Geoscience and Remote Sensing Letters},
  volume={6},
  number={4},
  pages={662--665},
  year={2009},
  publisher={IEEE}
}

@article{arienzo2022full,
  title={Full-resolution quality assessment of pansharpening: Theoretical and hands-on approaches},
  author={Arienzo, Alberto and Vivone, Gemine and Garzelli, Andrea and Alparone, Luciano and Chanussot, Jocelyn},
  journal={IEEE Geoscience and Remote Sensing Magazine},
  volume={10},
  number={3},
  pages={168--201},
  year={2022},
  publisher={IEEE}
}

@article{wald1997fusion,
  title={Fusion of satellite images of different spatial resolutions: Assessing the quality of resulting images},
  author={Wald, Lucien and Ranchin, Thierry and Mangolini, Marc},
  journal={Photogrammetric engineering and remote sensing},
  volume={63},
  number={6},
  pages={691--699},
  year={1997}
}

@article{deng2022machine,
  title={Machine learning in pansharpening: A benchmark, from shallow to deep networks},
  author={Deng, Liang-Jian and Vivone, Gemine and Paoletti, Mercedes E and Scarpa, Giuseppe and He, Jiang and Zhang, Yongjun and Chanussot, Jocelyn and Plaza, Antonio},
  journal={IEEE Geoscience and Remote Sensing Magazine},
  volume={10},
  number={3},
  pages={279--315},
  year={2022},
  publisher={IEEE}
}

@inproceedings{chen2022spanconv,
  title={SpanConv: A New Convolution via Spanning Kernel Space for Lightweight Pansharpening.},
  author={Chen, Zhi-Xuan and Jin, Cheng and Zhang, Tian-Jing and Wu, Xiao and Deng, Liang-Jian},
  booktitle={IJCAI},
  pages={841--847},
  year={2022}
}

@article{vivone2014critical,
  title={A critical comparison among pansharpening algorithms},
  author={Vivone, Gemine and Alparone, Luciano and Chanussot, Jocelyn and Dalla Mura, Mauro and Garzelli, Andrea and Licciardi, Giorgio A and Restaino, Rocco and Wald, Lucien},
  journal={IEEE Transactions on Geoscience and Remote Sensing},
  volume={53},
  number={5},
  pages={2565--2586},
  year={2014},
  publisher={IEEE}
}

@article{Kingma2014AdamAM,
  title={Adam: A Method for Stochastic Optimization},
  author={Diederik P. Kingma and Jimmy Ba},
  journal={CoRR},
  year={2014},
  volume={abs/1412.6980}
}

@inproceedings{Yang2019CondConvCP,
  title={CondConv: Conditionally Parameterized Convolutions for Efficient Inference},
  author={Brandon Yang and Gabriel Bender and Quoc V. Le and Jiquan Ngiam},
  booktitle={Neural Information Processing Systems},
  year={2019}
}

@article{Bengio2013EstimatingOP,
  title={Estimating or Propagating Gradients Through Stochastic Neurons for Conditional Computation},
  author={Yoshua Bengio and Nicholas L{\'e}onard and Aaron C. Courville},
  journal={ArXiv},
  year={2013},
  volume={abs/1308.3432}
}
\end{document}